\documentclass{article}
\usepackage{spconf,amsmath,graphicx}

\usepackage{enumitem}
\setlist{nosep, leftmargin=14pt}

\usepackage{mwe} 
\usepackage{algorithmic}
\usepackage{algorithm}
\usepackage[outdir=./]{epstopdf}
\usepackage{amssymb}
\usepackage{kantlipsum} 
\usepackage{mwe} 
\usepackage[noadjust]{cite}
\usepackage{multirow}
\usepackage{graphicx}
\usepackage{adjustbox}
\usepackage{subcaption}
\usepackage{amsmath}
\usepackage{etoolbox}
\usepackage{booktabs}
\usepackage{siunitx}
\usepackage{hyperref}

\patchcmd{\thebibliography}
  {\settowidth}
  {\setlength{\itemsep}{2pt plus 10pt}\settowidth}
  {}{}
\apptocmd{\thebibliography}
  {\small}
  {}{}
\title{Automated Classification of Cell Shapes: A Comparative Evaluation of Shape Descriptors}
\name{Valentina Vadori$^{1}$ \ Antonella Peruffo$^2$ \ Jean-Marie Graïc$^{2}$ \ Livio Finos$^{3}$ \ Enrico Grisan$^{1}$}
\address{$^{1}$London South Bank University, School of Engineering, United Kingdom
\\ $^{2}$University of Padova, Dept. of Comparative Biomedicine \& Food Science, Italy
\\ $^{3}$University of Padova, Dept. of Statistical Sciences, Italy}
\begin{document}
%
\maketitle
\begin{abstract}
This study addresses the challenge of classifying cell shapes from noisy contours, such as those obtained through cell instance segmentation of histological images. We assess the performance of various features for shape classification, including Elliptical Fourier Descriptors, curvature features, and lower-dimensional representations. 
Using an annotated synthetic dataset of noisy contours, we identify the most suitable shape descriptors and apply them to a set of real images for qualitative analysis. Our aim is to provide a comprehensive evaluation of descriptors for classifying cell shapes, which can support cell type identification and tissue characterization—critical tasks in both biological research and histopathological assessments. 
\end{abstract}

\section{Introduction}
\label{sec:intro}
Cell morphology is an essential factor in the study, diagnosis, prognosis, and treatment of diseases, since morphological abnormalities can signal pathological conditions \cite{alizadeh2020cellular,wu2020single}. Cell shape (i.e., the outline of a cell) is a critical component of cell morphology, closely related to cellular function and behavior. In both biological research and clinical settings, automated cell shape classification—when integrated with molecular and physiological data, along with other morphological characteristics—can aid in identifying cell types \cite{zeng2022cell,dance2024cell, corain2020multi} and quantitatively characterizing tissue \textit{cytoarchitecture}. \cite{amunts2015architectonic, graic2023cytoarchitectureal, graic2024age}.


%

\begin{figure}%
\centering
\includegraphics[width=0.9\columnwidth]{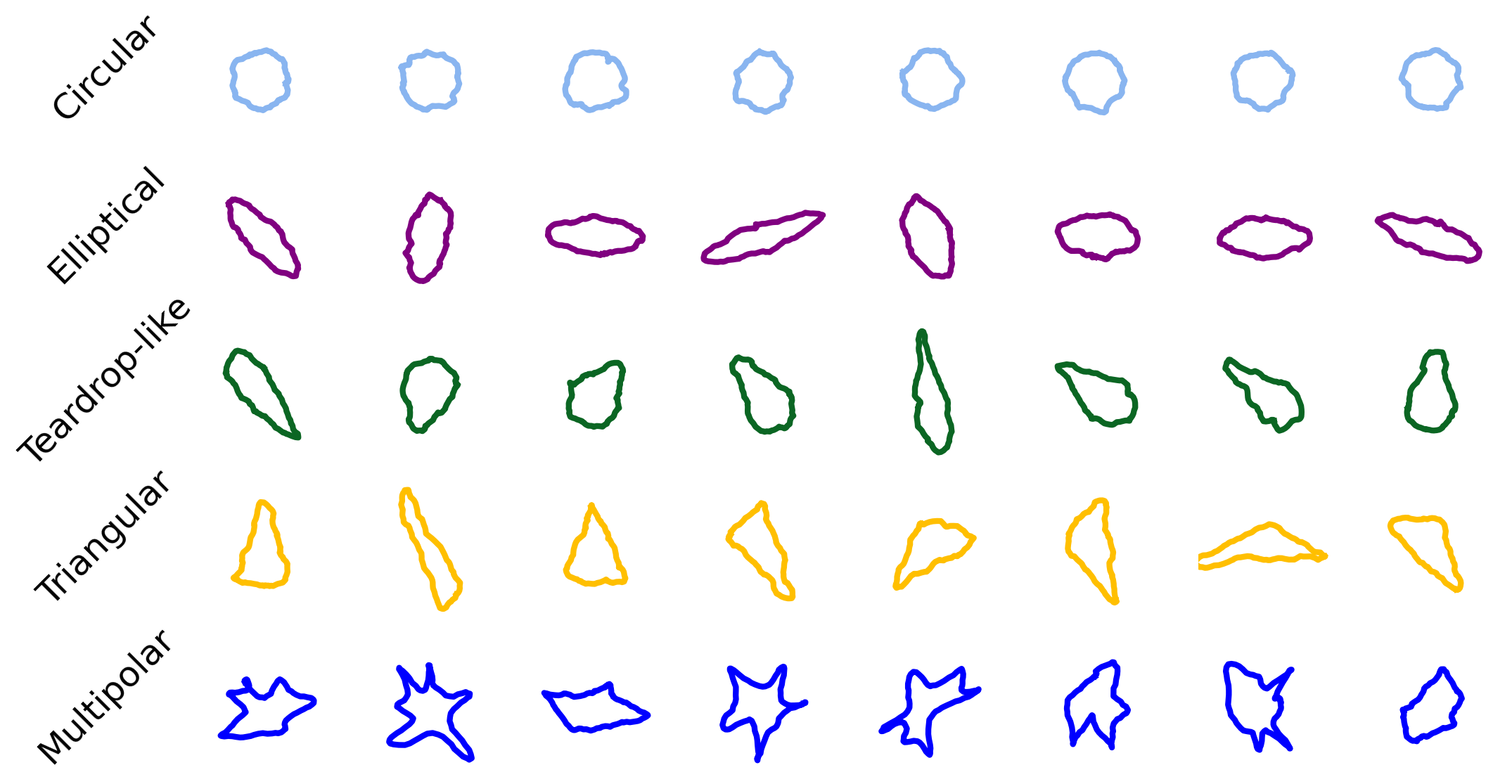}%
\caption{Noisy cell shapes synthetized for model training.}%
\label{fig:shapeclasses}%
\end{figure}

\begin{figure}%
\centering
\includegraphics[width=0.9\columnwidth]{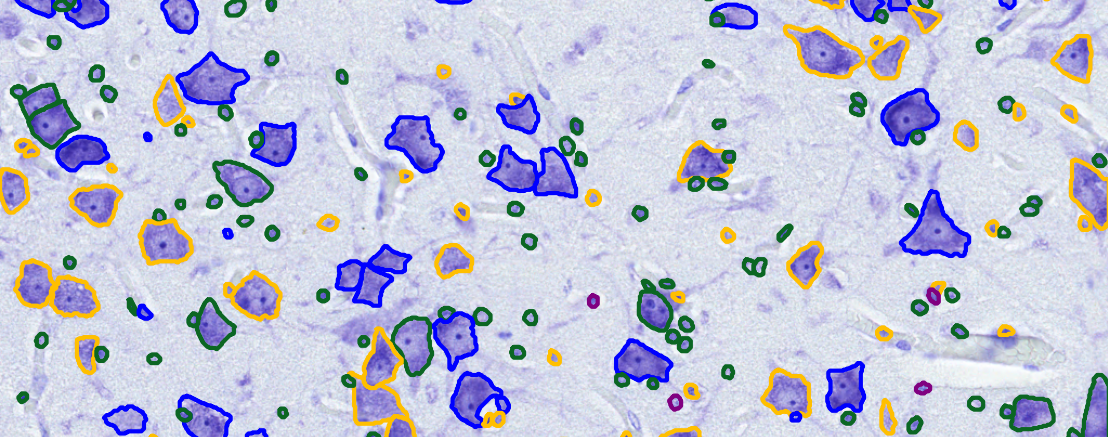}%
\caption{Qualitative results on real data.}%
\label{fig:example}%
\end{figure}

In this study, we focus on the challenge of classifying cells extracted from digitized histological brain images into distinct shape classes. We consider five shape classes, as illustrated in Fig. \ref{fig:shapeclasses}: circular, elliptical/spindle-like, teardrop-like/unipolar, triangular/pyramidal, and irregular/multipolar. These classes represent the typical shapes of neuronal and glial cell bodies \cite{kandel2000principles} observable in histological images stained with the Nissl method (as shown in Fig. \ref{fig:example}), which is widely regarded as the most effective technique for highlighting cell body morphology in tissue sections \cite{garcia2016distinction}.

Scalar features such as axis ratio and solidity have been used to characterize cell shapes. However, these features are often insufficient for consistently and accurately discriminating between shapes. While more complex solutions are available (cf. Section \ref{sec:related}), to our knowledge, no comprehensive, quantitative comparison of shape descriptors for cell shape classification has been conducted, and this study seeks to begin addressing that gap. As outlined in Section \ref{sec:methods}, we extract multiple shape descriptors to characterize cell contours, followed by classification and feature importance evaluation. 
The feature set that best performs on a synthetic dataset of 100k noisy contours is tested on CytoDArk0 (an open dataset of Nissl-stained histological images of the mammalian brain) for qualitative validation, as discussed in Section \ref{sec:results}. 

\section{Related Work}


The most common method for characterizing cell shape involves using scalar descriptors, such as circularity (the ratio of the area of a shape to the area of a circle with the same perimeter), solidity (the ratio of the area to the convex hull area), aspect ratio and axis ratio \cite{da2018shape}. Tools like CellProfiler and ImageJ can extract a wide range of these features for downstream analysis. However, scalar shape features may have limitations in capturing the complexity of cell morphology \cite{phillip2021robust}. Alternatively, dimensionality reduction techniques, such as Principal Component Analysis (PCA) \cite{pincus2007comparison} or autoencoders \cite{burgess2024orientation}, can compress the shape data into an optimized set of values that retain most of the morphological information. Another approach involves using transform-based techniques, including Fourier or Wavelet Transform, and Zernike Polynomials, to represent the shape with a set of coefficients indicating the contribution of different basis functions \cite{da2018shape}. 

Once the shape representations are established, either supervised or unsupervised machine learning techniques can be applied to assign cells to shape classes.
\label{sec:related}
\section{Methods}
\label{sec:methods}
\begin{figure*}[htbp]
\centering
\begin{subfigure}{0.43\columnwidth}{ 
\centering
    \includegraphics[height=1.1\columnwidth]{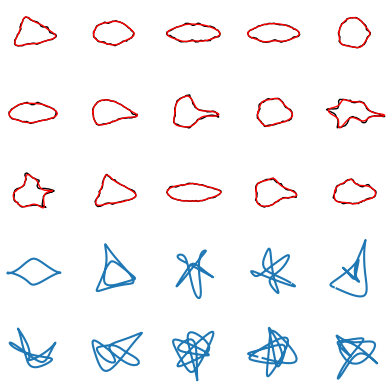}%
    \vspace{-0.1cm}}
    \caption{PCA}%
    \label{subfigb}%
\end{subfigure}%
\hspace{0.04\textwidth}
\begin{subfigure}{1.01\columnwidth}{ 
\centering
    \includegraphics[width=1\columnwidth]{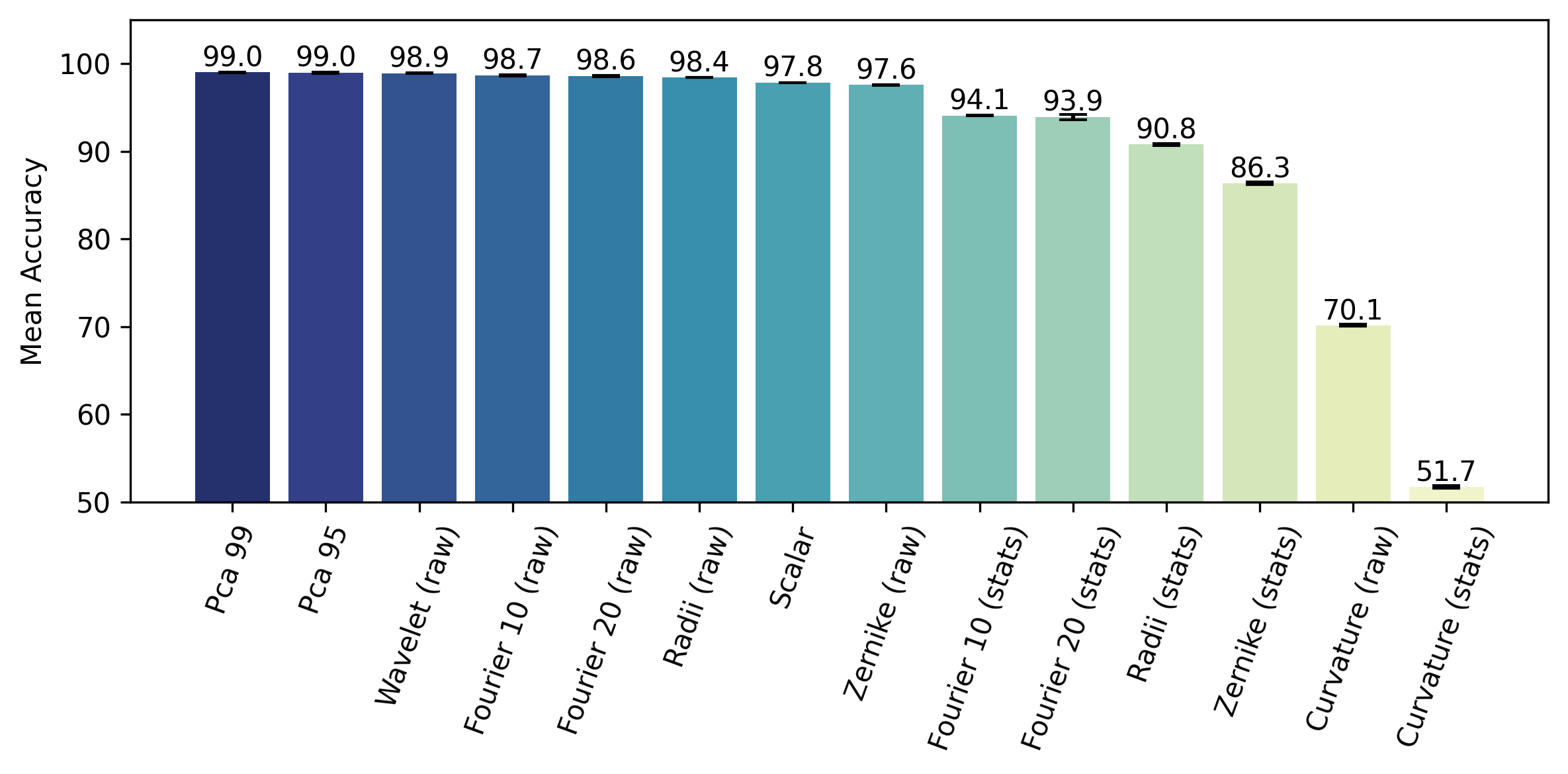}%
        \vspace{-0.33cm}}
    \caption{Classification performance}%
    \label{subfiga}%
\end{subfigure}%
\hspace{0.005\textwidth}
\begin{subfigure}{0.5\columnwidth}{ 
\centering
    \includegraphics[height=1\columnwidth]{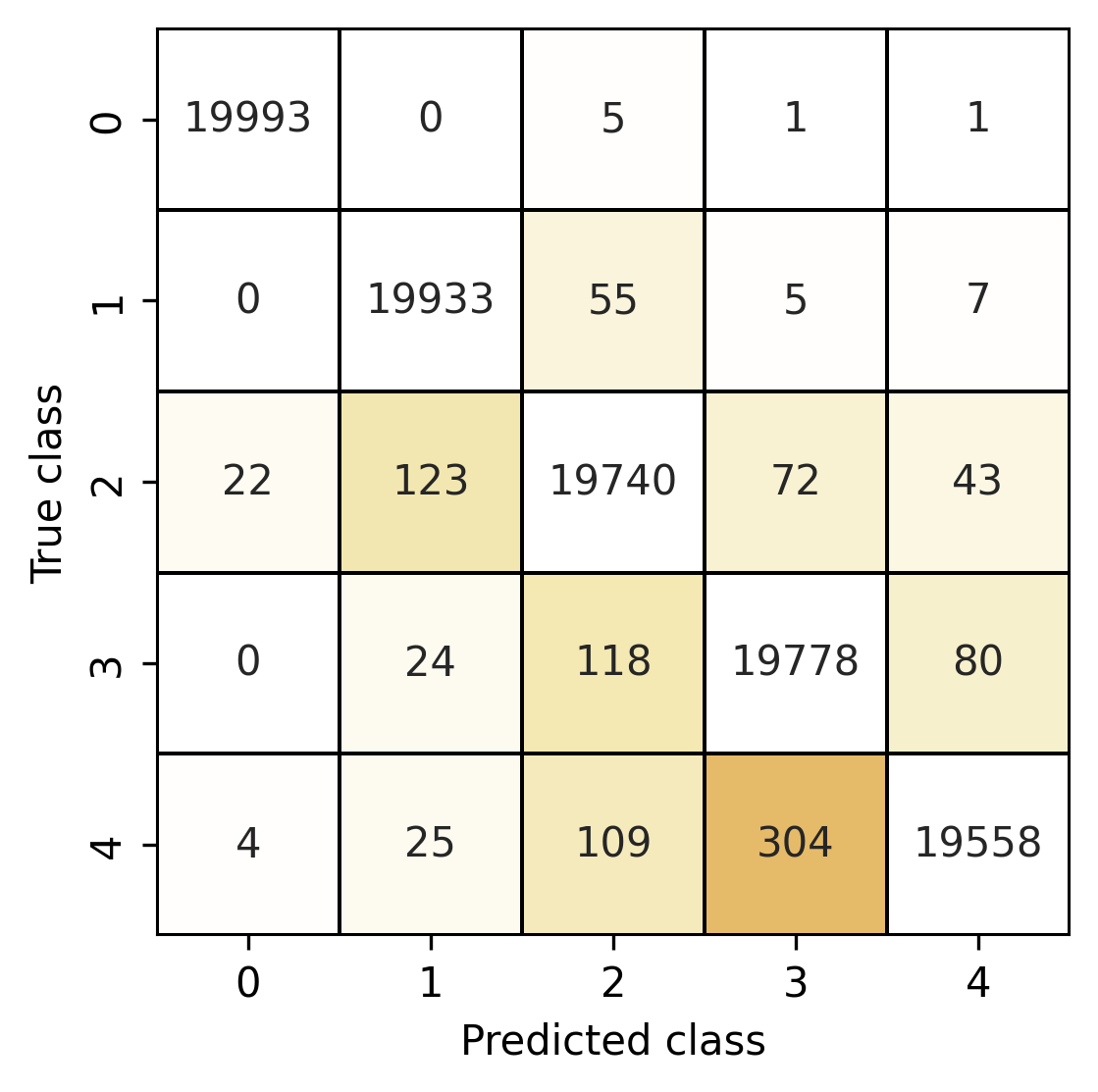}%
    \vspace{-0.33cm}}
    \caption{Confusion matrix} 
    \label{subfigc} 
\end{subfigure}

\vspace{0.3cm}

\begin{subfigure}{\textwidth}{
\centering
    \includegraphics[width=17.75cm]{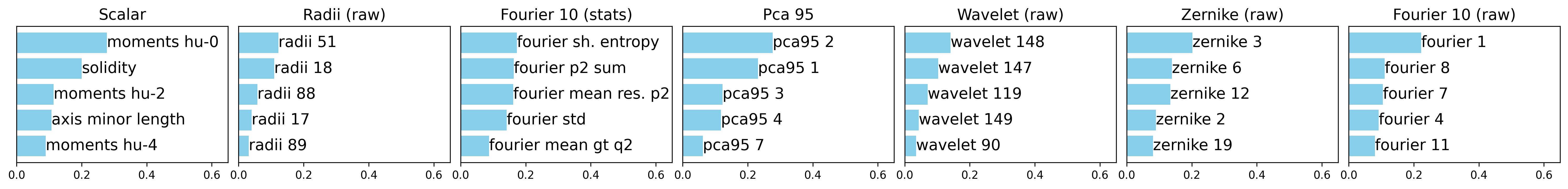}%
            \vspace{-0.25cm}}
    \caption{Feature Importance} 
    \label{subfigD} 
\end{subfigure}
\vspace{-0.6cm}
\caption{(a) Examples of registered contours (black) and reconstructed contours from PCA 95 (red, top 3 rows). First 10 shape modes from PCA 95 (cyan, bottom 2 rows). (b) Performance of contour descriptors for shape classification (c) Aggregated confusion matrix from the 5 folds of cross validation for PCA 99. (d) Top 5 important features for each feature set.}
\label{results:fig1}
\end{figure*}

\subsection{Contour Extraction and Pre-Processing}
We focus on classifying cell shapes in histological brain images. Deep learning models such as convolutional neural networks or, more recently, transformers, can be applied to identify the boundaries of individual cells within these images \cite{stringer2021cellpose, upschulte2022contour, horst2024cellvit, vadori2024cisca}. From the resulting label map, where each cell is assigned a unique integer, a contour-tracking algorithm extracts each cell's contour. This contour is represented as a $N \times 2$ matrix of 2D coordinates, with $N$ indicating the number of points.
Contours undergo preprocessing and Procrustes registration \cite{dryden2016statistical} to normalize and align them as follows. They are resampled to 100 points and reoriented to ensure a consistent clockwise orientation, aligned along their major axis, centered at $(0,0)$, rescaled to have the same area. Following this, the contours are iteratively rotated to minimize their distance from a reference shape, which is defined as the mean shape at the beginning of each iteration. This process continues until the change in the mean shape falls below a specified threshold, signifying that the alignment is complete.
\subsection{Feature Extraction}
\label{ssec:FeatureExtraction}
Feature extraction is carried out to capture the shape of the contours. Effective descriptors don’t require the ability to perfectly reconstruct shapes; instead, a strong set of descriptors should offer high discriminative power (reflected in classification performance) and efficiency, using as few descriptors as possible \cite{da2018shape}. We evaluate the following feature sets.

\textit{Scalar} features are the quantitative descriptors that encapsulate key characteristics of a shape in a simplified, one-dimensional representation \cite{bray2016cell,driscoll2012automated}. We consider the following 23: area of the bounding box, area of the convex hull, perimeter, major and minor axis lengths, axis ratio, eccentricity, extent,  solidity, circularity, roundness, maximum feret diameter,width and height of the minimal bounding box, width and height of the (standard) bounding box, Hu Moments.

\textit{Curvature} is the rate of change of the direction of the tangent vector \cite{da2018shape} and is assessed along contours by applying a spline fit to compute derivatives. The result is a vector of length $N$.

\textit{Radii} is the series of N distances between the shape centroid (0,0) and the shape boundary points \cite{da2018shape}. 

\textit{PCA} is applied to the coordinates of shapes after concatenating them into a single 1D vector \cite{pincus2007comparison,phillip2021robust}.  The principal components correspond to distinct shape modes (cf. Fig. \ref{subfigb}). The shapes are then encoded as a set of projection coefficients, which indicate how much each shape mode contributes to the representation of a given shape. We define \textit{PCA 99} and \textit{PCA 95} as the PCA that retain 99\% and 95\% of the variance, with coefficients of 32 and 14, respectively.

\textit{Elliptical Fourier Descriptors (EFD)}
Following \cite{kuhl1982elliptic}, a closed contour of order $M$ is a list of 2D coordinates 
represented using Fourier sine and cosine functions. 
The x-coordinate is parameterized by two series of coefficients $a = (a_0, a_1, \ldots, a_M)$ and $b = (b_1, \ldots, b_M)$, with $a_0$ determining the spatial offset of the contour. 
Likewise, 
the y-coordinate is parameterized by $c$ and $d$. The parameter vector $[a, b, c, d] \in \mathbb{R}^{4M + 2}$ fully determines the contour. 
The hyperparameter $M$ governs the smoothness of the contour; higher values of $M$ introduce higher frequency coefficients, allowing for more accurate approximations of object contours. 
We consider $\{a_i, b_i, c_i, d_i | 1 \leq i \leq M\}$, ignoring the offsets $(a_0, c_0)$, as \textit{Fourier} descriptors. We evaluate both order $M=10$ (\textit{Fourier 10}) and $M=20$ (\textit{Fourier 20}).

\textit{Wavelet}
When the wavelet transform \cite{antoine1997shape,yadav2007retrieval} is applied to the 1D signal \( r(t) \) representing the radii features, \( r(t) \) is decomposed using basis functions into a series of \textit{approximation} coefficients \( A[k] \), which capture the low-frequency, coarse features of the signal, and \textit{detail} coefficients \( D[k] \), which capture high-frequency, fine features. We applied the Haar wavelet transform and set \( K = 100 \) to represent each contour with the vector \( [A, D] \in \mathbb{R}^{2K} \) of \textit{Wavelet} descriptors.




\textit{Zernike moments}
are rotation-invariant image descriptors derived from Zernike polynomials, a collection of orthogonal polynomials defined on the unit disk \cite{da2018shape}. We computed a total of 25 moments on the binary mask that represents the contour, using a maximum polynomial degree of 8.

For descriptors like radii, curvature, Zernike moments, and EFD, we use the term \textit{raw} to indicate the original, unprocessed values of these descriptors. In contrast, \textit{stats} refers to a collection of 32 statistical metrics derived from the raw data, including mean, standard deviation, kurtosis, skewness, bending energy, and Shannon entropy.

%

\subsection{Classification}
\label{ssec:Classification} For classifying cell shapes we employed XGBoost \cite{chen2016xgboost}, which uses a gradient-boosted trees algorithm to create an ensemble of weak learners that sequentially correct previous errors. We applied 5-fold cross-validation, with each fold split into training (60\%), validation (20\%), and test sets (20\%). For hyperparameter optmization, we performed Random Search over the hyperparameter space on the training set. For each feature set, the variant that best performed on the validation set was selected for classification and feature importance evaluation on the test set. In XGBoost, key to the latter is the \textit{gain}, the reduction in the log loss function achieved when a feature is used to split branches.


Training was conducted on a synthetic, balanced dataset of 100k noisy contours, with examples displayed in Fig. \ref{fig:shapeclasses}. The code for generating the dataset and the dataset can be accessed at \url{https://github.com/vadori/cytoark}.

\section{Results}
\label{sec:results}
Fig. \ref{results:fig1} illustrates the classification performance of the shape descriptors under consideration, measured by mean accuracy.

The highest-performing methods are PCA 99 and 95, with 99.0\% accuracy. Wavelet (raw) and Fourier 10 (raw) are close behind with 98.9\% and 98.7\%, respectively. Scalar features score lower but still show strong performance. In general, stats-based descriptors underperform relative to their raw counterparts, indicating that raw representations retain the discriminative information needed for accurate classification. Curvature features scored the lowest accuracies.

The confusion matrix indicates that the majority of misclassifications occur with class 4 (Multipolar) being incorrectly predicted as class 3 (Triangular). Other common errors include class 2 (Teardrop-like) being predicted as 1 (Elliptical), and class 3 or 4 as 2. 
It is evident that for PCA 95, the most significant features are primarily the first coefficients. In contrast, for the Fourier (raw) method, intermediate frequencies hold greater importance. Notably, three Hu moments rank among the top five scalar features.

We used the Cytodark0 dataset \cite{vadori2024cytodark0, vadori2024cisca} and the best model trained on PCA 95 features for qualitative evaluation. As shown in Fig. \ref{fig:example}, most multipolar and triangular cells (blue) were effectively classified. Some small cells were identified as teardrop-like shapes, even though we expect larger cells to be teardrop-like. This mainly occurs because the model does not account for scale, and pointy contour parts can be exaggerated during rescaling. No round cells were detected, and this might be due to the confusing presence of circular shapes in the synthetic teardrop-like and elliptical classes.
\section{Conclusion}
PCA outperformed other methods for classifying cell shapes, followed by wavelet and EFD. Our findings provide support to the results in \cite{pincus2007comparison}, where the authors evaluated the contour reconstruction fidelity and interpretability of shape descriptors and
 favoured PCA.  For training, we used a large, balanced dataset, which provided an advantage for learning and resulted in very high accuracies. The framework outlined here can be applied to evaluate the robustness of descriptors in more challenging scenarios involving imbalanced datasets and increased noise levels. Ultimately, this will aid in the development of accurate and robust cell shape classification methods, which can enhance biological and medical initiatives that depend on cellular morphology characterization.

\vfill
\pagebreak

\bibliographystyle{IEEEtran}
\bibliography{strings,refs}

\begin{thebibliography}{10}
\providecommand{\url}[1]{#1}
\csname url@samestyle\endcsname
\providecommand{\newblock}{\relax}
\providecommand{\bibinfo}[2]{#2}
\providecommand{\BIBentrySTDinterwordspacing}{\spaceskip=0pt\relax}
\providecommand{\BIBentryALTinterwordstretchfactor}{4}
\providecommand{\BIBentryALTinterwordspacing}{\spaceskip=\fontdimen2\font plus
\BIBentryALTinterwordstretchfactor\fontdimen3\font minus
  \fontdimen4\font\relax}
\providecommand{\BIBforeignlanguage}[2]{{%
\expandafter\ifx\csname l@#1\endcsname\relax
\typeout{** WARNING: IEEEtran.bst: No hyphenation pattern has been}%
\typeout{** loaded for the language `#1'. Using the pattern for}%
\typeout{** the default language instead.}%
\else
\language=\csname l@#1\endcsname
\fi
#2}}
\providecommand{\BIBdecl}{\relax}
\BIBdecl

\bibitem{alizadeh2020cellular}
E.~Alizadeh, J.~Castle, A.~Quirk, C.~D. Taylor, W.~Xu, and A.~Prasad,
  ``Cellular morphological features are predictive markers of cancer cell
  state,'' \emph{Computers in biology and medicine}, vol. 126, p. 104044, 2020.

\bibitem{wu2020single}
P.-H. Wu, D.~M. Gilkes, J.~M. Phillip, A.~Narkar, T.~W.-T. Cheng, J.~Marchand,
  M.-H. Lee, R.~Li, and D.~Wirtz, ``Single-cell morphology encodes metastatic
  potential,'' \emph{Science advances}, vol.~6, no.~4, p. eaaw6938, 2020.

\bibitem{zeng2022cell}
H.~Zeng, ``What is a cell type and how to define it?'' \emph{Cell}, vol. 185,
  no.~15, pp. 2739--2755, 2022.

\bibitem{dance2024cell}
A.~Dance, ``What is a cell type, really? the quest to categorize life's myriad
  forms.'' \emph{Nature}, vol. 633, no. 8031, pp. 754--756, 2024.

\bibitem{corain2020multi}
L.~Corain, E.~Grisan, J.-M. Gra{\"\i}c, R.~Carvajal-Schiaffino, B.~Cozzi, and
  A.~Peruffo, ``Multi-aspect testing and ranking inference to quantify
  dimorphism in the cytoarchitecture of cerebellum of male, female and intersex
  individuals: a model applied to bovine brains,'' \emph{Brain Structure and
  Function}, vol. 225, no.~9, pp. 2669--2688, 2020.

\bibitem{amunts2015architectonic}
K.~Amunts and K.~Zilles, ``Architectonic mapping of the human brain beyond
  brodmann,'' \emph{Neuron}, vol.~88, no.~6, pp. 1086--1107, 2015.

\bibitem{graic2023cytoarchitectureal}
J.-M. Gra{\"\i}c, L.~Finos, V.~Vadori, B.~Cozzi, R.~Luisetto, T.~Gerussi,
  M.~Gatto, A.~Doria, E.~Grisan, L.~Corain \emph{et~al.}, ``Cytoarchitectureal
  changes in hippocampal subregions of the nzb/w f1 mouse model of lupus,''
  \emph{Brain, Behavior, \& Immunity-Health}, vol.~32, p. 100662, 2023.

\bibitem{graic2024age}
J.-M. Gra{\"\i}c, L.~Corain, L.~Finos, V.~Vadori, E.~Grisan, T.~Gerussi,
  K.~Orekhova, C.~Centelleghe, B.~Cozzi, and A.~Peruffo, ``Age-related changes
  in the primary auditory cortex of newborn, adults and aging bottlenose
  dolphins (tursiops truncatus) are located in the upper cortical layers,''
  \emph{Frontiers in Neuroanatomy}, vol.~17, p. 1330384, 2024.

\bibitem{kandel2000principles}
E.~R. Kandel, J.~H. Schwartz, T.~M. Jessell, S.~Siegelbaum, A.~J. Hudspeth,
  S.~Mack \emph{et~al.}, \emph{Principles of neural science}.\hskip 1em plus
  0.5em minus 0.4em\relax McGraw-hill New York, 2000, vol.~4.

\bibitem{garcia2016distinction}
M.~{\'A}. Garc{\'\i}a-Cabezas, Y.~J. John, H.~Barbas, and B.~Zikopoulos,
  ``Distinction of neurons, glia and endothelial cells in the cerebral cortex:
  an algorithm based on cytological features,'' \emph{Frontiers in
  Neuroanatomy}, vol.~10, p. 107, 2016.

\bibitem{da2018shape}
L.~da~Fona~Costa and R.~M. Cesar~Jr, \emph{Shape classification and analysis:
  theory and practice}.\hskip 1em plus 0.5em minus 0.4em\relax Crc Press, 2018.

\bibitem{phillip2021robust}
J.~M. Phillip, K.-S. Han, W.-C. Chen, D.~Wirtz, and P.-H. Wu, ``A robust
  unsupervised machine-learning method to quantify the morphological
  heterogeneity of cells and nuclei,'' \emph{Nature protocols}, vol.~16, no.~2,
  pp. 754--774, 2021.

\bibitem{pincus2007comparison}
Z.~Pincus and J.~Theriot, ``Comparison of quantitative methods for cell-shape
  analysis,'' \emph{Journal of microscopy}, vol. 227, no.~2, pp. 140--156,
  2007.

\bibitem{burgess2024orientation}
J.~Burgess, J.~J. Nirschl, M.-C. Zanellati, A.~Lozano, S.~Cohen, and
  S.~Yeung-Levy, ``Orientation-invariant autoencoders learn robust
  representations for shape profiling of cells and organelles,'' \emph{Nature
  Communications}, vol.~15, no.~1, p. 1022, 2024.

\bibitem{stringer2021cellpose}
C.~Stringer, T.~Wang, M.~Michaelos, and M.~Pachitariu, ``Cellpose: a generalist
  algorithm for cellular segmentation,'' \emph{Nature Methods}, vol.~18, no.~1,
  pp. 100--106, 2021.

\bibitem{upschulte2022contour}
E.~Upschulte, S.~Harmeling, K.~Amunts, and T.~Dickscheid, ``Contour proposal
  networks for biomedical instance segmentation,'' \emph{Medical image
  analysis}, vol.~77, p. 102371, 2022.

\bibitem{horst2024cellvit}
F.~H{\"o}rst, M.~Rempe, L.~Heine, C.~Seibold, J.~Keyl, G.~Baldini, S.~Ugurel,
  J.~Siveke, B.~Gr{\"u}nwald, J.~Egger \emph{et~al.}, ``Cellvit: Vision
  transformers for precise cell segmentation and classification,''
  \emph{Medical Image Analysis}, vol.~94, p. 103143, 2024.

\bibitem{vadori2024cisca}
V.~Vadori, J.-M. Graïc, A.~Peruffo, G.~Vadori, L.~Finos, and E.~Grisan,
  ``Cisca and cytodark0: a cell instance segmentation and classification method
  for histo(patho)logical image analyses and a new, open, nissl-stained dataset
  for brain cytoarchitecture studies,'' \emph{arXiv e-prints}, pp. arXiv--2409,
  2024.

\bibitem{dryden2016statistical}
I.~L. Dryden and K.~V. Mardia, \emph{Statistical shape analysis: with
  applications in R}.\hskip 1em plus 0.5em minus 0.4em\relax John Wiley \&
  Sons, 2016.

\bibitem{bray2016cell}
M.-A. Bray, S.~Singh, H.~Han, C.~T. Davis, B.~Borgeson, C.~Hartland,
  M.~Kost-Alimova, S.~M. Gustafsdottir, C.~C. Gibson, and A.~E. Carpenter,
  ``Cell painting, a high-content image-based assay for morphological profiling
  using multiplexed fluorescent dyes,'' \emph{Nature protocols}, vol.~11,
  no.~9, pp. 1757--1774, 2016.

\bibitem{driscoll2012automated}
M.~K. Driscoll, J.~L. Albanese, Z.-M. Xiong, M.~Mailman, W.~Losert, and K.~Cao,
  ``Automated image analysis of nuclear shape: what can we learn from a
  prematurely aged cell?'' \emph{Aging (Albany NY)}, vol.~4, no.~2, p. 119,
  2012.

\bibitem{kuhl1982elliptic}
F.~P. Kuhl and C.~R. Giardina, ``Elliptic fourier features of a closed
  contour,'' \emph{Computer graphics and image processing}, vol.~18, no.~3, pp.
  236--258, 1982.

\bibitem{antoine1997shape}
J.-P. Antoine, D.~Barachea, R.~M. Cesar~Jr, and L.~da~Fontoura~Costa, ``Shape
  characterization with the wavelet transform,'' \emph{Signal Processing},
  vol.~62, no.~3, pp. 265--290, 1997.

\bibitem{yadav2007retrieval}
R.~B. Yadav, N.~K. Nishchal, A.~K. Gupta, and V.~K. Rastogi, ``Retrieval and
  classification of shape-based objects using fourier, generic fourier, and
  wavelet-fourier descriptors technique: A comparative study,'' \emph{Optics
  and Lasers in engineering}, vol.~45, no.~6, pp. 695--708, 2007.

\bibitem{chen2016xgboost}
T.~Chen and C.~Guestrin, ``Xgboost: A scalable tree boosting system,'' in
  \emph{Proceedings of the 22nd Acm Aigkdd International Conference on
  Knowledge Discovery and Data Mining}, 2016, pp. 785--794.

\bibitem{vadori2024cytodark0}
\BIBentryALTinterwordspacing
V.~Vadori, J.-M. Graïc, A.~Peruffo, G.~Vadori, L.~Finos, and E.~Grisan,
  ``Cytodark0,'' Sep. 2024. [Online]. Available:
  \url{https://doi.org/10.5281/zenodo.13694738}
\BIBentrySTDinterwordspacing

\end{thebibliography}

\end{document}